\documentclass{article}

 \usepackage[final]{neurips_2021}

\usepackage[utf8]{inputenc} 
\usepackage[T1]{fontenc}    
\usepackage{hyperref}       
\usepackage{url}            
\usepackage{booktabs}       
\usepackage{amsfonts}       
\usepackage{nicefrac}       
\usepackage{microtype}      
\usepackage{xcolor}         

\usepackage{amsmath} 
\usepackage[pdftex]{graphicx} 
\usepackage{subcaption} 
\usepackage{bm}

\title{Compositional Modeling of Nonlinear Dynamical Systems with ODE-based Random Features}

\author{%
  Thomas M. McDonald \\
  Department of Computer Science\\
  University of Sheffield\\
  \texttt{tmmcdonald1@sheffield.ac.uk} \\
  \And
  Mauricio A.~\'Alvarez \\
  Department of Computer Science\\
  University of Sheffield\\
  \texttt{mauricio.alvarez@sheffield.ac.uk} \\
}

\begin{document}

\maketitle

\begin{abstract}
Effectively modeling phenomena present in highly nonlinear dynamical systems whilst also accurately quantifying uncertainty is a challenging task, which often requires problem-specific techniques. We present a novel, domain-agnostic approach to tackling this problem, using compositions of physics-informed random features, derived from ordinary differential equations. The architecture of our model leverages recent advances in approximate inference for deep Gaussian processes, such as layer-wise weight-space approximations which allow us to incorporate random Fourier features, and stochastic variational inference for approximate Bayesian inference. We provide evidence that our model is capable of capturing highly nonlinear behaviour in real-world multivariate time series data. In addition, we find that our approach achieves comparable performance to a number of other probabilistic models on benchmark regression tasks.
\end{abstract}

\section{Introduction}
Dynamical systems are ubiquitous across the natural sciences, with many physical and biological processes being driven on a fundamental level by differential equations. Inferring ordinary differential equation (ODE) parameters using observational data from such systems is an active area of research \citep{meeds2019efficient, ghosh2021variational}, however, in particularly complex systems it is often infeasible to characterise all of the individual processes present and the interactions between them. Rather than attempt to fully describe a complex system, latent force models (LFMs) \citep{alvarez2009latent} specify a simplified mechanistic model of the system which captures salient features of the dynamics present. This leads to a model which is able to readily extrapolate beyond the training input space, thereby retaining one of the key advantages of mechanistic modeling over purely data-driven techniques.

Modeling \textit{nonlinear} dynamical systems presents an additional challenge, and whilst nonlinear differential equations have been incorporated into LFMs \citep{ward2020black}, shallow models such as LFMs are generally less capable than deep models of modeling the non-stationarities often present in nonlinear systems. Deep model architectures have greater representational power than shallow models as a result of their hierarchical structure, which typically consists of compositions of functions \citep{lecun2015deep}. Deep probabilistic models such as deep Gaussian processes (DGPs) \citep{damianou2013deep} are able to leverage this representational power, with the additional advantage of being able to accurately quantify uncertainty. 

In this paper, we aim to model nonlinear dynamics by constructing a DGP from compositions of physics-informed random Fourier features, with the motivation behind this approach being that many real-world systems are compositional hierarchies \citep{lecun2015deep}. Whilst a number of recent works have incorporated random Fourier features into deep models \citep{cutajar2017random, mehrkanoon2018deep}, the only work we are aware of which does so in the context of a physics-informed DGP is that of \citet{lorenzi2018constraining}. The authors impose physical structure on a DGP with random features by constraining the function values within the model as a means of performing ODE parameter inference. Rather than following the approach of \citet{lorenzi2018constraining} and placing constraints on function values whilst using an exponentiated quadratic (EQ) kernel, we instead integrate a physics-informed prior into the structure of the DGP by utilising a kernel based on an ODE.

\begin{figure}
\centering
\begin{subfigure}{0.99\linewidth}
  \centering
  \includegraphics[width=0.99\linewidth]{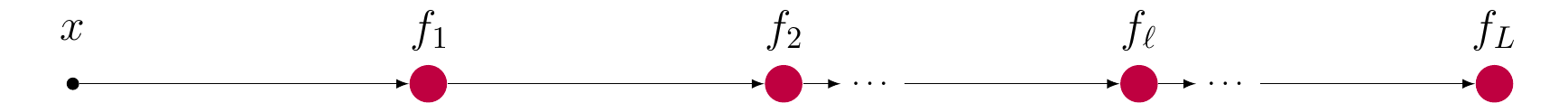}
  \caption{Deep Gaussian process (DGP)}
  \label{fig:dlfm_diagram_chain_a}
\end{subfigure}
\begin{subfigure}{0.99\linewidth}
  \centering
  \includegraphics[width=0.99\linewidth]{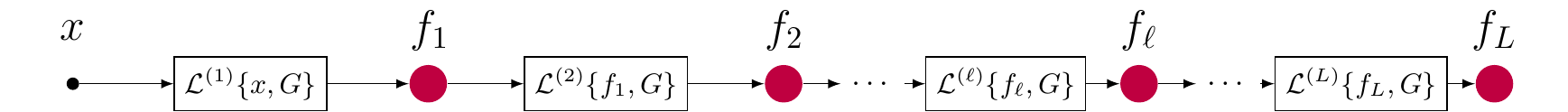}
  \caption{Deep latent force model (DLFM)}
  \label{fig:dlfm_diagram_chain_b}
\end{subfigure}
  \caption{A conceptual explanation of how our proposed DLFM differs from a DGP. At each layer, we perform the operation $\mathcal{L}^{(\ell)}\{x, G\} = \int_0^x G^{(\ell)}(x-\tau)u(\tau)d\tau$, where $G$ is the Green's function corresponding to an ODE, and $u(\cdot)$ represents an exponentiated quadratic GP prior. For example, the second operation in the model shown above would take the form, $\mathcal{L}^{(2)}\{f_1, G\} = \int_0^{f_1} G^{(2)}(f_1-\tau)u(\tau)d\tau$.}
  \label{fig:dlfm_diagram_chain}
\end{figure}

Our main contribution in this paper is the introduction of a novel approach to incorporating physical structure into a deep probabilistic model, whilst providing a sound quantification of uncertainty. We achieve this through derivation of physics-informed random Fourier features via the convolution of an EQ GP prior with the Green's function associated with a first order ODE. These features are then incorporated into each layer of a DGP, as shown in Figure \ref{fig:dlfm_diagram_chain}. To ensure the scalability of our model to large datasets, stochastic variational inference is employed as a method for approximate Bayesian inference. We provide experimental evidence that our modeling framework is capable of capturing highly nonlinear dynamics effectively in both toy examples and real world data, whilst also being applicable to more general regression problems.

\section{Background}
In this section, we review the theory behind using random Fourier features for deep Gaussian processes. We also review the idea of building covariance functions for GPs that encode the dynamics of ODEs and how to compute those covariances using random Fourier features. 

\subsection{Deep Gaussian Processes with Random Feature Expansions}\label{section:rff:dgp}

Gaussian processes (GPs) are non-parametric probabilistic models which
offer a great degree of flexibility in terms of the data they can
represent, however this flexibility is not unbounded. The hierarchical
networks of GPs now widely known as deep Gaussian processes (DGPs)
were first formalised by \citet{damianou2013deep}, with the motivating
factor behind their creation being the ability of deep networks to reuse features and allow for higher levels of abstraction in said features \citep{bengio2013representation}, which results in such models having more representational power and flexibility than shallow models such as GPs. DGPs
are effectively a composite function, where the input to the model is
transformed to the output by being passed through a series of latent
mappings (i.e. multivariate GPs). If we consider a supervised learning
problem with inputs denoted by $\mathbf{X} = \{\mathbf{x}_n\}^N_{n=1}$
and targets denoted by $\mathbf{y} = \{y_n\}^N_{n=1}$, we can write
the analytic form of the marginal likelihood for a DGP with $N_h$
hidden layers as
$p(\mathbf{y}|\mathbf{X}, \mathbf{\theta}) = \int
p(\mathbf{y}|\mathbf{F}^{(N_h)})p(\mathbf{F}^{(N_h)}|\mathbf{F}^{(N_h
  - 1)}, \mathbf{\theta}^{(N_h - 1)}) \ ... \
p(\mathbf{F}^{(1)}|\mathbf{X},
\mathbf{\theta}^{(0)})d\mathbf{F}^{(N_h)} \ ... \ d\mathbf{F}^{(1)}$, where $\mathbf{F}^{(\ell)}$ and $\mathbf{\theta}^{(\ell)}$ represent
the latent values and covariance parameters respectively at the
$\ell$-th layer, where $\ell = 0, ... , N_h$. However, due to the
need to propagate densities through non-linear GP covariance functions
within the model, this integral is intractable
\citep{damianou2015deep}. As this precludes us from employing exact
Bayesian inference in these models, various techniques for approximate inference have been applied to DGPs in recent years, with most approaches broadly based upon either variational inference \citep{salimbeni2017doubly, salimbeni2019deep, yu2019implicit} or Monte Carlo methods \citep{havasi2018inference}.

\citet{cutajar2017random} outline an alternative approach to tackling this
problem, which involves replacing the GPs present at each layer of the
network with their two layer weight-space approximation, 
forming a Bayesian neural network which acts as an approximation to a
DGP. If the $\ell$-th layer of such a DGP (consisting of zero-mean GPs
with exponentiated quadratic kernels) receives an input
$\mathbf{F}^{(\ell)}$ (where $\mathbf{F}^{(0)} = \mathbf{X}$), the
random features for this layer are denoted by 
$\mathbf{\Phi}^{(\ell)} \in \mathbb{R}^{N \times 2N^{(\ell)}_{RF}}$ 
(where $N^{(\ell)}_{RF}$ denotes the number of random features used), and are given by,
\begin{align} \label{eq:dgp_features}
\mathbf{\Phi}^{(\ell)} = \sqrt[]{\frac{(\sigma^2)^{(\ell)}}{N_{RF}^{(\ell)}}} \left[ \cos(\mathbf{F}^{(\ell)}\mathbf{\Omega}^{(\ell)}), \sin(\mathbf{F}^{(\ell)}\mathbf{\Omega}^{(\ell)}) \right] ,
\end{align} 
where $(\sigma^2)^{(\ell)}$ is the marginal variance kernel hyperparameter, $N_{RF}^{(\ell)}$ is 
the number of random features used and $\mathbf{\Omega}^{(\ell)} \in \mathbb{R}^{D_{F^{(\ell)}} \times
  N_{RF}^{(\ell)}}$ is the matrix of spectral frequencies used to
determine the random features. This matrix is assigned a prior
$p(\Omega^{(\ell)}_d) = \mathcal{N}(\Omega^{(\ell)}_d \mid 0,
{\left(l^{(\ell)}\right)}^{-2})$ where $l^{(\ell)}$ is the lengthscale kernel hyperparameter,
$D_{F^{(\ell)}}$ is the number of GPs within the layer, and $d = 1, ..., D_{F^{(\ell)}}$. The
random features then undergo the linear transformation, $\mathbf{F}^{(\ell + 1)} = \mathbf{\Phi}^{(\ell)} \mathbf{W}^{(\ell)}$, where $\mathbf{W}^{(\ell)} \in \mathbb{R}^{2N_{RF}^{(\ell)} \times D_{F^{(\ell + 1)}}}$ is a weight matrix with each column assigned a standard normal prior. 
Training is achieved via \textit{stochastic variational inference}, which involves establishing a tractable lower bound for the marginal likelihood and 
optimising said bound with respect to the mean and variance of the variational distributions over the weights and spectral frequencies across all 
layers of the network. The bound is also optimised with respect to the kernel hyperparameters across all layers.

\subsection{Random Feature Approximations for Latent Force Models}\label{sec:RFF:LFM}
In this paper, we take a similar approach to \citet{cutajar2017random}, but we utilise physically-inspired random features. Physics-inspired covariance functions for GPs have been proposed by \citep{alvarez2009latent} under the name of latent force models, and more recently, the authors of \citet{guarnizo2018fast} studied the construction of random features related to such covariance functions.

Latent force models (LFMs) \citep{alvarez2009latent} are GPs which incorporate a physically-inspired kernel function, which typically encodes the behaviour described by a specific form of differential equation. Instead of taking a fully mechanistic approach and specifying all the interactions within a physical system, the kernel describes a simplified form of the system in which the behaviour is determined by $Q$ latent forces. Given an input data-point $t$ and a set of $D$ output variables $\{f_d(t)\}^D_{d=1}$, a LFM expresses each output as $f_d(t) = \sum_{q=1}^Q S_{d, q} \int_{0}^t G_d(t-\tau)u_q(\tau)d\tau$,
 where $G_d(\cdot)$ represents the Green's function for a certain form of linear differential equation, $u_q(t)\sim \mathcal{GP}(0, k_q(t,t'))$ represents the GP prior over the the $q$-th latent force, and $S_{d,q}$ is a sensitivity parameter weighting the influence of the $q$-th latent force on the $d$-th output. The general expression for the covariance of a LFM is given by $k_{f_d, f_{d'}}(t,t') = \sum_{q=1}^Q S_{d,q} S_{d',q} \int^t_0 G_d(t-\tau) \int^{t'}_0 G_{d'}(t'-\tau') k_q(\tau, \tau') d\tau' d\tau$.

Typically, an EQ form is assumed for the kernel governing the latent forces, $k_q(\cdot)$. Due to the linear nature of the convolution used to compute $f_d(t)$, the outputs can also be described by a GP. Exact inference in LFMs is tractable, however as with GPs, it scales with $\mathcal{O}(N^3)$ complexity \citep{Rasmussen06}. Furthermore, computing the double integration for $k_{f_d, f_{d'}}(t,t')$ leads to terms that include the numerical approximation of complex functions which burden the computation of the kernel function. By providing a random Fourier feature representation for the EQ kernel $k_q(t,t')$, \citet{guarnizo2018fast} were able to reduce this cubic dependence on the number of data-points to a linear dependence. This representation arises from Bochner's theorem, which states,
\begin{align}
k_q(\tau, \tau') = e^{\frac{-(\tau-\tau')^2}{\ell_q^2}} &=\int p(\omega)e^{j(\tau-\tau')\omega} d\omega \approx \frac{1}{N_{RF}}  \sum^{N_{RF}}_{s=1} e^{j\omega_s \tau} e^{-j\omega_s \tau'} , \label{eq:bochner2}
\end{align}
where $\ell_q$ is the lengthscale of the EQ kernel, $N_{RF}$ is the number of random features used in the approximation and $\omega_s \sim p(\omega) = \mathcal{N}(\omega | 0, 2/\ell_q^2)$. Substituting this form of the EQ kernel into $k_{f_d, f_{d'}}(t,t')$ leads to a fast approximation to the LFM covariance $k_{f_d, f_{d'}}(t,t') \approx \sum_{q=1}^Q \frac{S_{d,q}S_{d',q}}{N_{RF}} \left[ \sum_{s=1}^{N_{RF}} \phi_d(t, \theta_d, \omega_s) \phi_{d'}^*(t', \theta_{d'}, \omega_s)  \right]$,
where 
\begin{align}
\phi_d(t, \theta_d, \omega) = \int_0^t G_d(t-\tau)e^{j\omega \tau} d\tau, \label{eq:RFRF}
\end{align}
with $\theta_d$ representing the Green's function parameters. When the Green's function is a real function, $\phi_{d'}^*(t', \theta_{d'}, \omega) = \phi_{d'}(t', \theta_{d'}, -\omega)$. \citet{guarnizo2018fast} refer to $\phi_d(t, \theta_d, \omega)$ as \emph{random Fourier response features} (RFRFs).

\section{Deep Latent Force Models}
In this section, we present the deep latent force model (DLFM), a novel approach to incorporating physically-inspired prior beliefs into a deep Gaussian process. Firstly, we will outline the model architecture, before discussing our approach to training the model via stochastic variational inference.

\subsection{Model Formulation}
Rather than deriving the random features $\mathbf{\Phi^{(\ell)}}$ within the DGP from an EQ kernel \citep{cutajar2017random}, we instead populate this matrix with features derived from an LFM kernel. The exact form of the features derived is dependent on the Green's function used, which in turn depends on the form of differential equation whose characteristics we wish to encode within the model. \citet{guarnizo2018fast} derived a number of different forms corresponding to various differential equations, with the simplest case being that of a first order ordinary differential equation (ODE) of the form,$\frac{df(t)}{dt} + \gamma f(t) = \sum^Q_{q=1} S_q u_q(t)$, where $\gamma$ is a decay parameter associated with the ODE and $S_q$ is a sensitivity parameter which weights the effect of each latent force. For simplicity of exposition, we assume $D=1$. The Green's function associated to this ODE has the form $G(x) = e^{-\gamma x}$. By using equation \eqref{eq:RFRF}, the RFRFs associated with this ODE follow as
\begin{align}
\phi(t, \gamma, \omega_s) & = \frac{e^{j\omega_s t} - e^{-\gamma t}}{\gamma + j\omega_s} ,  \label{eq:ode1_feature}
\end{align}
where compared to the expression \eqref{eq:RFRF}, the parameter $\theta$ of the Green's function corresponds to the decay parameter $\gamma$.
From here on, we redefine the spectral frequencies as $\omega_{q, s}$ to emphasise the fact that the values sampled are dependent on the lengthscale of the 
latent force by way of the prior, $\omega_{q,s} \sim \mathcal{N}(\omega | 0, 2/\ell_q^2)$.
We can collect all of the random features corresponding to the $q$-th latent force into a single vector, $\bm{\phi}^c_q(t, \gamma, \bm{\omega}_q) = \sqrt{S_q^2/N_{RF}}[\phi(t,
  \gamma, \omega_{q, 1}), \cdots , \phi(t, \gamma, \omega_{q, N_{RF}})]^\top \in \mathbb{C}^{N_{RF} \times 1}$, where $\bm{\omega}_q=\{\omega_{q,s}\}_{s=1}^{N_{RF}}$, $\mathbb{C}$ refers to the complex plane and the super index $c$ in $\bm{\phi}_q^c(\cdot)$ makes explicit that this vector contains complex-valued numbers. By including the random features corresponding to all $Q$ latent forces within the model, we obtain $\bm{\phi}^c(t, \gamma, \bm{\omega}) = [\left(\bm{\phi}^c_1(t, \gamma, \bm{\omega}_1\right)^{\top}, \cdots , \left(\bm{\phi}^c_Q(t, \gamma, \bm{\omega}_Q)\right)^{\top}]^\top \in \mathbb{C}^{QN_{RF} \times 1}$, with $\bm{\omega}=\{\bm{\omega}_q\}_{q=1}^Q$. These random features will be denoted as $\bm{\phi}^c_{LFM}(t, \gamma, \bm{\omega})$ to differentiate them from the features $\bm{\phi}(\cdot)$ computed from a generic EQ kernel.

\paragraph{Higher-dimensional inputs} Although the expression for $\bm{\phi}^c_{LFM}(t, \gamma,\bm{\omega})$ was obtained in the context of an ODE where the input is the time variable, in this paper we exploit this formalism to propose the use of these features even in the context of a generic supervised learning problem where the input is a potentially high-dimensional vector $\mathbf{x} = [x_1, x_2, \cdots, x_p]^\top\in\mathbb{R}^{p\times 1}$. As will be noticed later, such an extension is also necessary if we attempt to use such features at intermediate layers of the composition. Essentially, we compute a vector $\bm{\phi}^c_{LFM}(x_m, \gamma_m, \bm{\omega}_m)$ for each input dimension $x_{m}$ leading to a set of vectors  $\{\bm{\phi}^c_{LFM}(x_1, \gamma_1, \bm{\omega}_1),\cdots, \bm{\phi}^c_{LFM}(x_p, \gamma_p, \bm{\omega}_p)\}$. Notice that the samples $\bm{\omega}_m$ can also be different per input dimension, $x_m$. Although there are different ways in which these feature vectors can be combined, in this paper, we assume that the final random feature vector computed over the whole input vector $\mathbf{x}$ is given as $\bm{\phi}^c_{LFM}(\mathbf{x}, \bm{\gamma}, \bm{\Omega}) = \sum_{m=1}^p \bm{\phi}^c_{LFM}(x_m, \gamma_m, \bm{\omega}_m)$, where $\bm{\Omega}=\{\bm{\omega}_m\}_{m=1}^p$ and $\bm{\gamma}=\{\gamma_m\}_{m=1}^p$. An alternative to explore for future work involves expressing $\bm{\phi}^c_{LFM}(\mathbf{x}, \bm{\gamma}, \bm{\Omega})$ as $\bm{\phi}^c_{LFM}(\mathbf{x}, \bm{\gamma}, \bm{\Omega}) = \sum_{m=1}^p \alpha_m \bm{\phi}^c_{LFM}(x_m, \gamma_m, \bm{\omega}_m)$, with $\alpha_m\in \mathbb{R}$ a parameter that weights the contribution of each input feature differently. Although we allow each input dimension to have a different decay parameter $\gamma_m$ in the experiments in Section \ref{experiments}, for ease of notation we will assume that $\gamma_1=\gamma_2=\cdots=\gamma_p=\gamma$. For simplicity, we write $\bm{\phi}^c_{LFM}(\mathbf{x}, \gamma, \bm{\Omega})$. Therefore, $\bm{\phi}^c_{LFM}(\mathbf{x}, \gamma, \bm{\Omega})$ is a vector-valued function that maps from $\mathbb{R}^{p\times 1}$ to $\mathbb{C}^{QN_{RF} \times 1}$.

\paragraph{Real version of the RFRFs} Rather than working with the complex-value random features $\bm{\phi}^c_{LFM}(\mathbf{x}, \gamma, \bm{\Omega})$, we can work with their real-valued counterpart by using $\bm{\phi}_{LFM}(\mathbf{x}, \gamma, \bm{\Omega}) = [\left(\mathfrak{Re}\{\bm{\phi}^c_{LFM}(\mathbf{x}, \gamma, \bm{\Omega})\}\right)^\top , \left(\mathfrak{Im}\{\bm{\phi}^c_{LFM}(\mathbf{x}, \gamma, \bm{\Omega})\}\right)^\top]^{\top}\in \mathbb{R}^{2QN_{RF}\times 1}$ \citet{guarnizo2018fast}, where $\mathfrak{Re}(a)$ and $\mathfrak{Im}(a)$
take the real component and imaginary component of $a$, respectively. For an input matrix $\mathbf{X} = [\mathbf{x}_1, \cdots, \mathbf{x}_N]^{\top}\in \mathbb{R}^{N\times p}$, the matrix $\bm{\Phi}_{LFM}(\mathbf{X}, \gamma, \bm{\Omega}) = [\bm{\phi}_{LFM}(\mathbf{x}_1, \gamma, \bm{\Omega}), \cdots, \bm{\phi}_{LFM}(\mathbf{x}_N, \gamma, \bm{\Omega})]^{\top}\in \mathbb{R}^{N\times 2QN_{RF}}$.

\begin{figure}
\centering
\begin{subfigure}{0.99\linewidth}
  \centering
  \includegraphics[width=0.99\linewidth]{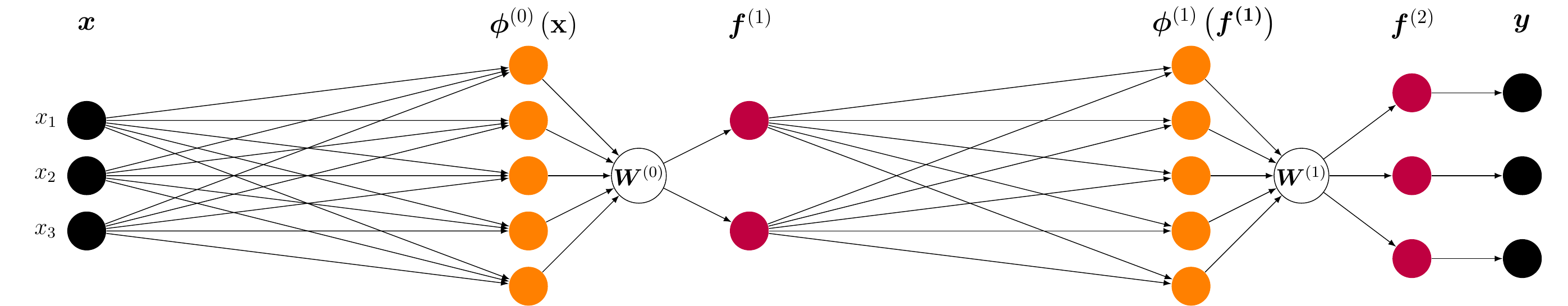}
  \caption{Deep Gaussian process (DGP)}
  \label{fig:dlfm_diagram_2a}
\end{subfigure}
\begin{subfigure}{0.99\linewidth}
  \centering
  \includegraphics[width=0.99\linewidth]{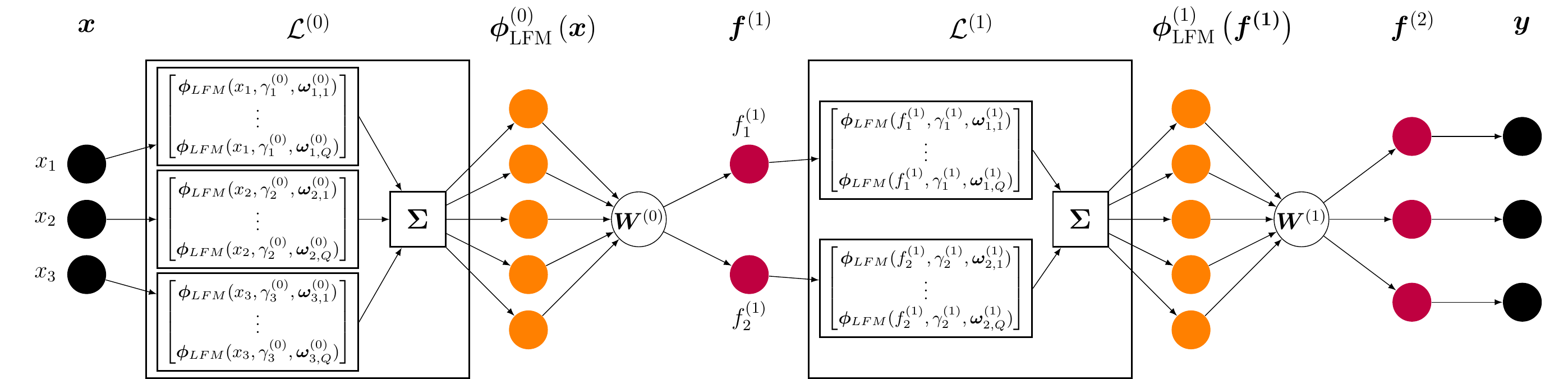}
  \caption{Deep latent force model (DLFM)}
  \label{fig:dlfm_diagram_2b}
\end{subfigure}
  \caption{An illustration of how our proposed model differs from a DGP with random feature expansions, with this example containing two layers. At each layer of the DLFM, for each input dimension, $N_{RF}$ random features of the form shown in \eqref{eq:ode1_feature} are computed for each of the $Q$ latent forces. The random feature vector $\bm{\phi}_{LFM}^{(\ell)}$ is then formed by taking the sum of these features across the input dimensions. This summation is shown in the Figure by the block containing $\Sigma$.}
  \label{fig:dlfm_diagram_2}
\end{figure}

\paragraph{Composition of RFRFs} We now use $\bm{\Phi}_{LFM}(\mathbf{X}, \gamma, \bm{\Omega})$ as a building block of a layered architecture of RFRFs. We write $\bm{\Phi}^{(\ell)}_{LFM}(\mathbf{F}^{(\ell)}, \gamma^{(\ell)}, \bm{\Omega}^{(\ell)})\in \mathbb{R}^{N\times 2Q^{(\ell)}N^{(\ell)}_{RF}}$ and follow a similar construction to the one described in
Section \ref{section:rff:dgp}, where $\mathbf{F}^{(\ell+1)} = \bm{\Phi}^{(\ell)}_{LFM}(\mathbf{F}^{(\ell)}, \gamma^{(\ell)}, \bm{\Omega}^{(\ell)})\mathbf{W}^{(\ell)}$, and $\mathbf{W}^{(\ell)}\in\mathbb{R}^{2Q^{(\ell)}N^{(\ell)}_{RF}\times D_{F^{(\ell+1)}}}$. As before, $\mathbf{F}^{(0)}=\mathbf{X}$. Figure \ref{fig:dlfm_diagram_2} is an example of this compositional architecture of RFRFs, which we refer to as a \emph{deep latent force model} (DLFM). When considering multiple-output problems, we allow extra flexibility to the decay parameters and lengthscales at the final ($L$-th) layer such that they vary not only by input dimension, but also by output. Mathematically, this corresponds to computing $\mathbf{F}_d^{(L)} = \bm{\Phi}^{(L-1)}_{LFM}(\mathbf{F}^{(\ell)}, \gamma_d^{(L-1)}, \bm{\Omega}_d^{(L-1)})\mathbf{W}_d^{(L-1)}$, where $d=1,...,D$.

\paragraph{Model interpretation} In a DGP, the layers of GPs successively warp the input space, propagating uncertainty through the model. The DLFM also performs a similar warping of the inputs, however,  in addition, the Green's function involved in the convolution which takes place within our model can be interpreted as a filter, which selectively filters out some of the frequencies of the GPs at each layer. Therefore, the DLFM is not only warping the intermediate inputs, but also filtering components in the latent GPs, altering their degree of smoothness.

\subsection{Variational Inference}
As previously mentioned, exact Bayesian inference is intractable for models such as ours, therefore in order to train the DLFM we employ stochastic variational inference \citep{hoffman2013stochastic}. For notational simplicity, let $\mathbf{W}$, $\mathbf{\Omega}$ and $\mathbf{\Theta}$ represent the collections of weight matrices, spectral frequencies and kernel hyperparameters respectively, across all layers of the model. We seek to optimise the variational distributions over $\mathbf{\Omega}$ and $\mathbf{W}$ whilst also optimising $\mathbf{\Theta}$, however we do not place a variational distribution over these hyperparameters. Our approach resembles the VAR-FIXED training strategy described by \citet{cutajar2017random} which involves reparameterizing $\Omega_{ij}^{(\ell)}$ such that $\Omega_{ij}^{(\ell)} = s_{ij}^{(\ell)} \epsilon_{rij}^{(\ell)} + m_{ij}^{(\ell)}$ , where $m_{ij}^{(\ell)}$ and $(s^2)_{ij}^{(\ell)}$ represent the means and variances associated with the variational distribution over $\Omega_{ij}^{(\ell)}$, and ensuring that the standard normal samples $\epsilon_{rij}^{(\ell)}$ are fixed throughout training rather than being resampled at each iteration. If we denote $\mathbf{\Psi} = \left\{\mathbf{W}, \mathbf{\Omega} \right\}$ and consider training inputs $\mathbf{X} \in \mathbb{R}^{N \times D_{in}}$ and outputs $\mathbf{y} \in \mathbb{R}^{N \times D_{out}}$, we can derive a tractable lower bound on the marginal likelihood using Jensen's inequality, which allows for minibatch training using a subset of $M$ observations from the training set of $N$ total observations. This bound, derived by \citet{cutajar2017random}, takes the form,
\begin{align}
\log[p(\mathbf{y} | \mathbf{X}, \mathbf{\Theta}] &= E_{q(\mathbf{\Psi})} \log[p(\mathbf{y} | \mathbf{X, \Psi, \Theta})] - \text{DKL}[q(\mathbf{\Psi}) || p(\mathbf{\Psi} | \mathbf{\Theta})] \\
&\approx \left[\frac{N}{M} \sum_{k \in \mathcal{I}_M} \frac{1}{N_{\text{MC}}} \sum^{N_{\text{MC}}}_{r=1} \log[p(\mathbf{y}_k | \mathbf{x}_k, \tilde{\mathbf{\Psi}}_r, \mathbf{\Theta})]\right] - \text{D}_{\text{KL}}[q(\mathbf{\Psi})||p(\mathbf{\Psi} | \mathbf{\Theta})] ,
\end{align}
where $\text{D}_{\text{KL}}$ denotes the KL divergence, $\tilde{\bm{\Psi}}_r \sim q(\bm{\Psi})$, the minibatch input space is denoted by $\mathcal{I}_M$ and $N_{\text{MC}}$ is the number of Monte Carlo samples used to estimate $E_{q(\bm{\Psi})}$. $q(\bm{\Psi})$ and $p(\bm{\Psi} | \bm{\Theta})$ denote the approximate variational distribution and the prior distribution over the parameters respectively, both of which are assumed to be Gaussian in nature. A full derivation of this bound and the expression for the KL divergence between two normal distributions are included in the supplemental material.

We mirror the approach of \citet{cutajar2017random} and \citet{KingmaW13} by reparameterizing the weights and spectral frequencies, which allows for stochastic optimisation of the means and variances of our distributions over $\bm{W}$ and $\bm{\Omega}$ via gradient descent techniques. Specifically, we use the AdamW optimizer \citep{loshchilov2018decoupled}, implemented in PyTorch, as empirically it led to superior performance compared to other alternatives such as conventional stochastic gradient descent.

\section{Related Work} \label{related}
As mentioned previously, the work of \citet{lorenzi2018constraining} also aims to incorporate physical structure into a deep probabilistic model, however the authors achieve this by constraining the dynamics of their DGP with random feature expansions, rather than specifying a physics-informed kernel as in our approach. Additionally, the model developed by \cite{lorenzi2018constraining} is primarily designed for tackling the problem of ODE parameter inference in situations where the form of the underlying ODEs driving the behaviour of a system are known. In contrast, our DLFM does not assume any knowledge of the differential equations governing the complex dynamical systems we study in Section \ref{experiments}, and thus we do not attempt to perform parameter inference; our primary aim is to construct a robust, physics-inspired model with extrapolation capabilities and the ability to quantify uncertainty in its predictions.

The work of \citet{mehrkanoon2018deep} is another example of a deep model in which each layer consists of a feature mapping computed using random Fourier features, followed by a linear transformation. Whilst this bears a similarity to the architecture of our model and to that of \citet{lorenzi2018constraining}, the authors only consider an EQ kernel for their feature mapping, with no consideration given to physically-inspired features or constraining function dynamics. \citet{wang2019deep} also propose a scalable deep probabilistic model capable of modeling non-linear functions (they specifically consider multivariate time series), but their approach relies on a deep neural network to model global behaviour, whilst relying on GPs at each layer to capture random effects and local variations specific to individual time series. \citet{zammit2021deep} is another example of recent work which aims to model nonstationary processes using a deep probabilistic architecture, with their primary focus being spatial warping. In \citet{duncker2018temporal}, the authors propose a nested GP model in which the inner GP warps the temporal input it is provided. If we consider a univariate time input, our model performs a similar process, with the weight-space GP layer warping the temporal input and passing the output into a second GP layer. However, unlike in \citet{duncker2018temporal}, our framework allows for an arbitrary number of warping transformations.

We wish to clarify that although our model is based on LFMs defined by convolutions of the form used to compute $f_d(t)$ in Section \ref{sec:RFF:LFM}, our model differs from the convolutional DGP construction outlined by \citet{dunlop2018deep}, $f^{(\ell + 1)}(t) = \int f^{(\ell)}(t-\tau)u^{(\ell + 1)}(\tau)d \tau$, which the authors argue results in trivial behaviour with increasing depth. Instead, our model takes the form, $f^{(\ell + 1)}(f^{(\ell)}) = \int_0^{f^{(\ell)}} G(f^{(\ell)} - \tau^{(\ell + 1)})u^{(\ell + 1)}(\tau^{(\ell + 1)}) d \tau^{(\ell + 1)}$, which is more akin to the compositional construction outlined by the \citet{dunlop2018deep}, just with a kernel which happens to involve a convolution.

\section{Experiments}
\label{experiments}
All of the experimental results in this section were obtained using a single node of a cluster, consisting of a 40 core Intel Xeon Gold 6138 CPU and a NVIDIA Tesla V100 SXM2 GPU with 32GB of RAM. Unless otherwise specified, all models in this section were implemented in pure PyTorch, trained using the AdamW optimizer with a learning rate of 0.01 and a batch size of 1000.\footnote{Our code is publicly available in the repository: \url{https://github.com/tomcdonald/Deep-LFM}. The code was also included within the supplemental material at the time of review.} The DLFMs and DGPs with random feature expansions \citep{cutajar2017random} tested all utilised a single hidden layer of dimensionality $D_{F^{(\ell)}} = 3$, 100 Monte Carlo samples and 100 random features per layer. Further details regarding experimental setups are provided in the supplemental material, alongside additional experimental results.

\subsection{Compositional System Toy Problem}
\begin{figure}
\centering
\begin{subfigure}{0.325\linewidth}
  \centering
  \includegraphics[width=0.99\linewidth]{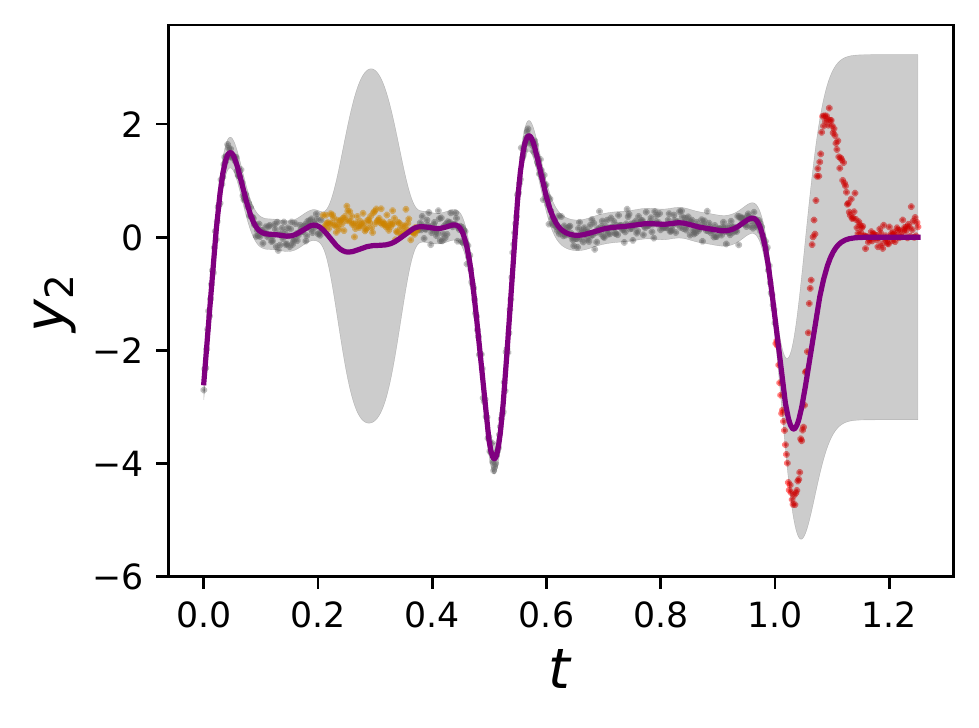}
  \caption{GP}
\end{subfigure}
\begin{subfigure}{0.325\linewidth}
  \centering
  \includegraphics[width=0.99\linewidth]{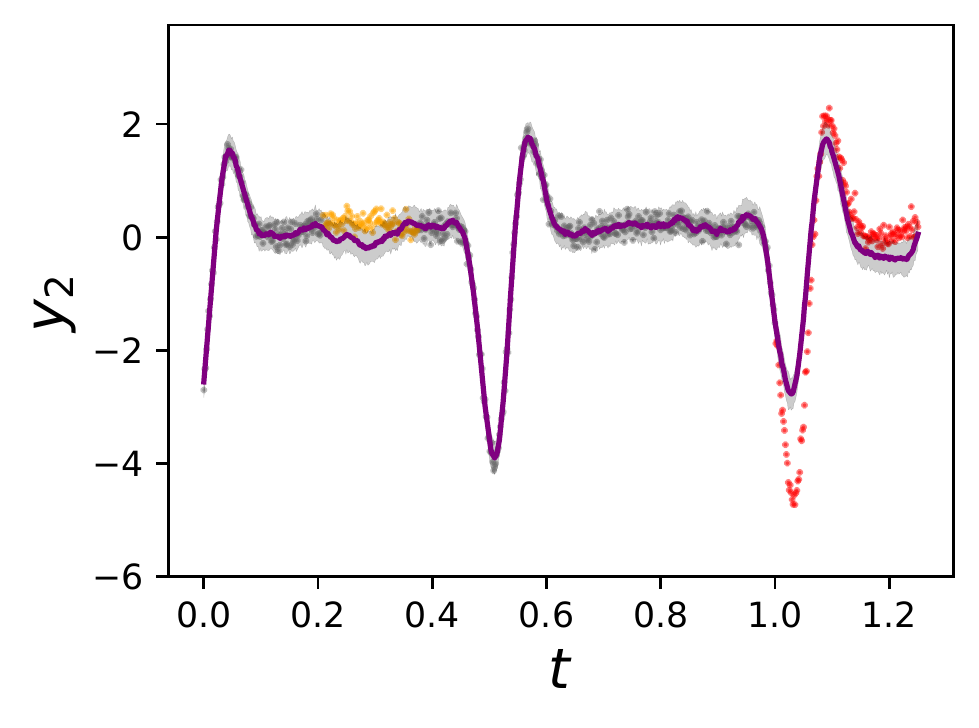}
  \caption{DGP}
  \label{fig:toy_dgp}
\end{subfigure}
\begin{subfigure}{0.325\linewidth}
  \centering
  \includegraphics[width=0.99\linewidth]{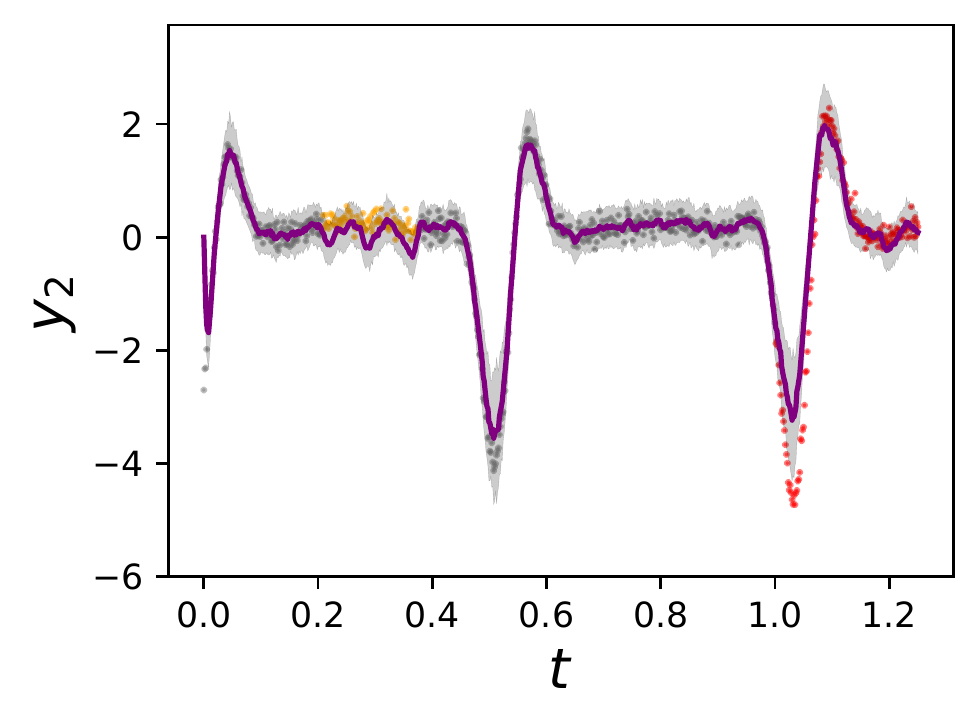}
  \caption{DLFM}
  \label{fig:toy_dlfm}
\end{subfigure}
  \caption{Three model fits to data from our toy dynamical system with the latent function $u(t)=\cos(0.5 t) + 6\sin(3 t)$. Grey, orange and red data-points represent training data, interpolation test data and extrapolation test data respectively. The purple curves represent the predictive means, and the shaded areas represent $\pm 2\sigma$.}
  \label{fig:toy_figure}
\end{figure}

Firstly, to verify that our model is capable of modeling the compositional systems which its architecture resembles, 
we train the model on the vector-valued inputs $t$ and noise-corrupted outputs $y_2(t)=f_2(t) + \epsilon$ of a system characterised 
by $f_1(t) = \int_0^t G_1(t-\tau)u(\tau)d\tau$ and $f_2(f_1) = \int_0^{f_1} G_2(f_1 - \tau^\prime) u(\tau^\prime) d\tau^\prime$,
where $G_1(\cdot)$ and $G_2(\cdot)$ represent the Green's functions corresponding to the first order ODEs 
which these two integrals represent the solutions to, and $\epsilon \sim \mathcal{N}(0, 0.04)$. We evaluate an exact GP, a DLFM (with a single latent 
force and 100 random features) and a DGP with random features, with a batch size equal to the size of the training set. 
The latent function is sinusoidal in nature, resulting in a signal consisting of pulses with varying amplitude. From the model 
predictions shown in Figure \ref{fig:toy_figure}, we see that all three models capture the behaviour of the system well in the 
training regime, but only the deep models are able to extrapolate. The DLFM and DGP have similar normalised mean squared errors (NMSE) 
on the extrapolation test set of 1.5 and 1.4 respectively, however, the DLFM appears to provide a more realistic quantification of predictive uncertainty, which is supported by the fact that the mean negative log likelihood (MNLL) evaluated on 
the extrapolation test set for the DLFM was 2.4, whereas for the DGP it was 3.7.

\subsection{PhysioNet Multivariate Time Series}
\label{ecg_extrapolation}
\begin{figure}
  \centering
  \includegraphics[width=0.95\linewidth]{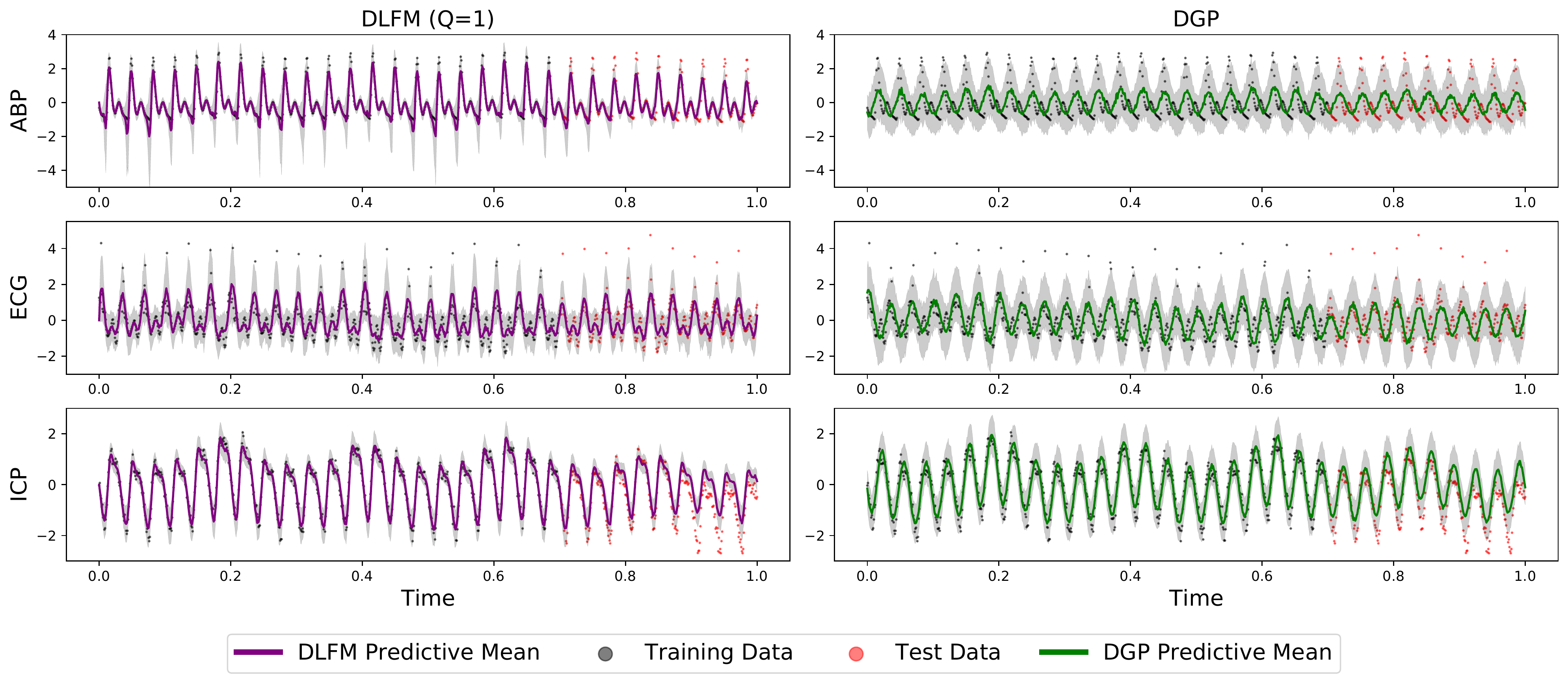}
  \caption{Predictions generated for each of the three outputs within the CHARIS dataset from a DLFM with one latent force (in the left-hand column), and a DGP with an exponentiated quadratic kernel (in the right-hand column). The shaded grey areas in each plot represent $\pm 2\sigma$.}
  \label{fig:ECG_figure}
\end{figure}

\begin{table}
  \caption{Extrapolation test set results for each output within the CHARIS dataset, with standard error in brackets.}
  \setlength{\tabcolsep}{2.5pt}
  \label{CHARIS_table}
  \centering
  \begin{tabular}{ccccccc}
    \toprule &
    \multicolumn{2}{c}{ABP} & \multicolumn{2}{c}{ECG} & \multicolumn{2}{c}{ICP} \\
    \cmidrule(r){2-3} \cmidrule(r){4-5} \cmidrule(r){6-7}
         & NMSE & MNLL & NMSE & MNLL & NMSE & MNLL \\
    \midrule
    DLFM ($Q=1$) & \textbf{0.39 (0.05)} & \textbf{0.75 (0.07)} & \textbf{0.45 (0.04)} & \textbf{0.99 (0.05)} & 0.36 (0.04) & 0.90 (0.06) \\
    DLFM ($Q=2$) & 0.42 (0.07) & 0.77 (0.08) & 1.58 (0.98) & 1.16 (0.16) & \textbf{0.32 (0.06)} & \textbf{0.88 (0.11)} \\
    DGP-EQ & 0.73 (0.002) & 1.21 (0.001) & 0.66 (0.01) & 1.22 (0.01) & 0.44 (0.01) & 1.00 (0.01) \\
	DGP-ARC & 1.01 (0.001) & 2.60 (0.02) & 1.03 (0.001) & 2.73 (0.04) & 1.36 (0.003) & 3.25 (0.02) \\
	VAR-GP & 1.01 (0.0002) & 26.90 (9.02) & 1.02 (0.001) & 4.98 (0.10) & 1.33 (0.003) & 15.50 (6.48) \\
	DNN & 1.04 (0.01) & N/A & 1.08 (0.02) & N/A & 2.41 (0.09) & N/A \\
	LFM-RFF & 1.00 (0.003) & N/A & 1.11 (0.02) & N/A & 1.02 (0.01) & N/A \\
    \bottomrule
  \end{tabular}
\end{table}

To assess the ability of our model to capture highly nonlinear behaviour, we evaluate its performance on a subset of the CHARIS dataset (ODC-BY 1.0 License) \citep{kim2016trending}, which can be found on the PhysioNet data repository \citep{goldberger2000physiobank}. The data available for each patient consists of an electrocardiogram (ECG), alongside arterial blood pressure (ABP) and intracranial pressure (ICP) measurements; all three of these signals are sampled at regular intervals and are nonlinear in nature. We use a subset of this data consisting of the first 1000 time steps for a single patient. We test two variations of the DLFM in this section, one with two latent forces and 50 random features per force ($Q=2$), and one with a single latent force and 100 random features ($Q=1$). To clarify, both of these models contain 100 random features per layer, but in the $Q=2$ case, 50 features are derived from two distinct latent forces. We compare these models to two DGPs with random features, employing EQ (DGP-EQ) and arccosine (DGP-ARC) kernels, as well as a deep neural network (DNN) with a single hidden layer of dimensionality 300, ReLU activations, and a dropout rate of 0.5. In addition to deep models, we also evaluate the performance of a heterogeneous multi-output GP trained using stochastic variational inference (VAR-GP) \citep{moreno2019heterogeneous} and a shallow LFM with random Fourier features derived from the same first order ODE as our model (LFM-RFF) \citep{guarnizo2018fast}, consisting of two latent forces and 20 random features. Experimental results comparing the ability of each of these models to impute values within the training-input space are presented in the supplemental material, however we focus here on the more challenging task of extrapolating beyond the training input-space by training the aforementioned models on the first 700 observations and withholding the remaining 300 as a test set.

The results in Table \ref{CHARIS_table} show that our model outperforms the other techniques tested, with the single latent force variant in particular converging to a significantly lower NMSE and MNLL than the other approaches across all three outputs. From these results we can conclude that the improved performance we see is a result of the ODE-based random features in the DLFM allowing the model to extrapolate accurately beyond the training input-space, rather than purely due to the additional flexibility afforded by allowing an increased number of lengthscales per layer, as the $Q=1$ model contains the same number of lengthscales per layer as the DGP. We also find that the deep models tested consistently outperform the shallow models. From Figure \ref{fig:ECG_figure}, we can see that the DLFM is more able to accurately extrapolate the nonlinear dynamics present in the system than the DGP, especially the variations in ABP.

\subsection{UCI Regression Benchmarks}
\begin{figure}
  \centering
  \includegraphics[width=0.95\linewidth]{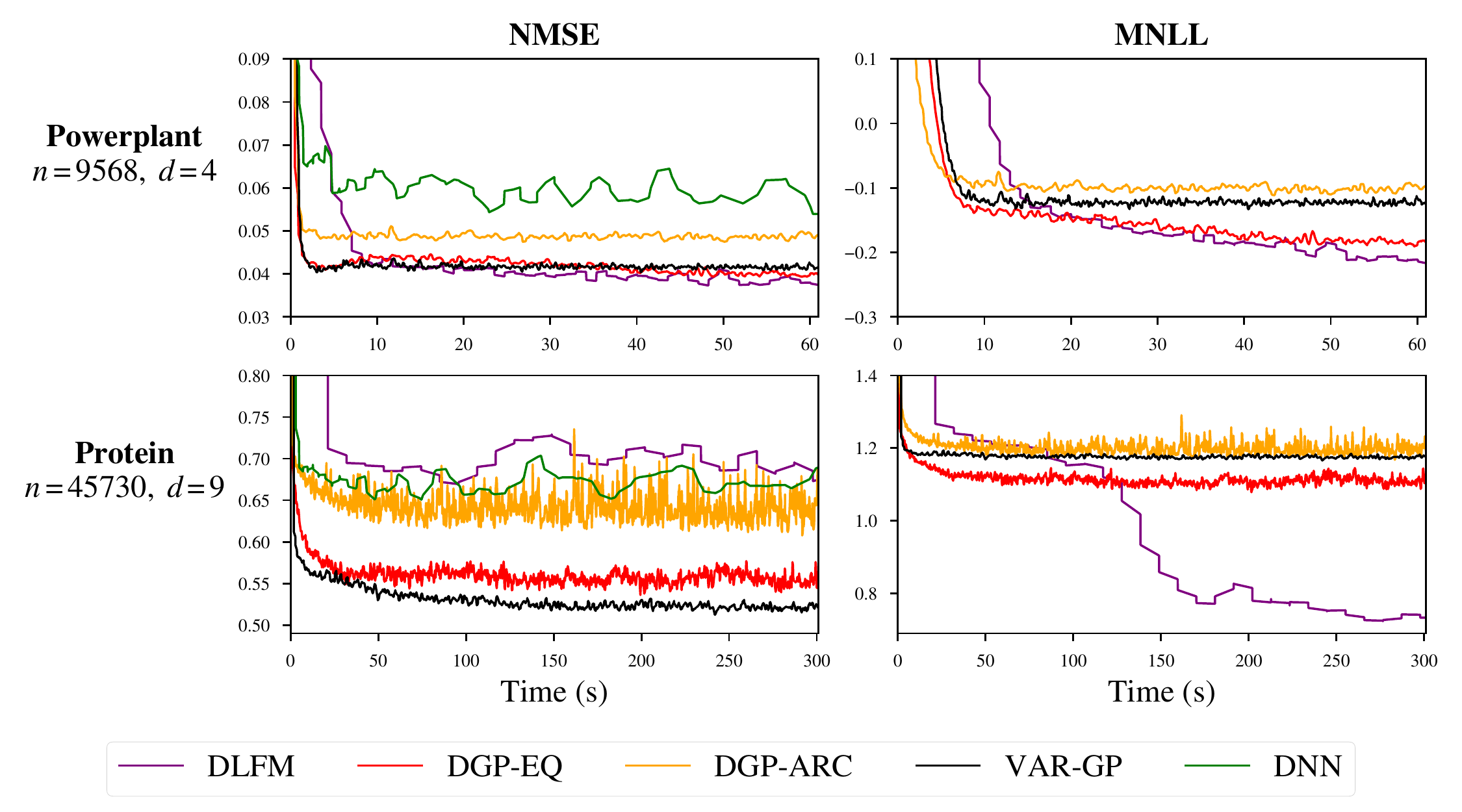}
  \caption{Progression of validation set metrics on the UCI benchmarks, averaged over three folds.}
  \label{fig:UCI_figure}
\end{figure}

Finally, we also evaluated the performance of the model on two regression datasets from the UCI Machine Learning Repository \citep{Dua2019}, 'Powerplant' and 'Protein', which both consist of a multivariate input and a univariate target. For both of these experiments, we evaluate the same DLFM, DGP and DNN architectures as for the PhysioNet experiment, alongside a multi-task variational GP (VAR-GP) \citep{hensman2015scalable} implemented using GPyTorch \citep{gardner2018gpytorch}. Achieving high performance on benchmark regression datasets was not the primary motivation behind this work, however from the results shown in Figure \ref{fig:UCI_figure} and Table \ref{UCI_table}, we find that the DLFM exhibits comparable performance to the models described above. Most notably, our model consistently converges to a superior mean negative log likelihood (MNLL), both on the validation set during training and also on the held-out test set.

\begin{table}
  \caption{Test set results on the UCI benchmarks, with standard error in brackets.}
  \label{UCI_table}
  \centering
  \begin{tabular}{ccccc}
    \toprule &
    \multicolumn{2}{c}{Powerplant} & \multicolumn{2}{c}{Protein} \\
    \cmidrule(r){2-3} \cmidrule(r){4-5}
         & NMSE & MNLL & NMSE & MNLL \\
    \midrule
    DLFM & \textbf{0.0557 (0.0005)} & \textbf{- 0.017 (0.005)} & 0.698 (0.004) & \textbf{0.754 (0.006)} \\
    DGP-EQ & 0.0581 (0.0002) & \ \ 0.016 (0.003) & 0.553 (0.005) & 1.101 (0.002)  \\
    DGP-ARC & 0.0619 (0.0002) & \ \ 0.039 (0.002) & 0.644 (0.027) & 1.203 (0.008)  \\
    VAR-GP & 0.0576 (0.0001) & \ \ 0.050 (0.001)  & \textbf{0.520 (0.005)} & 1.174 (0.002) \\
    DNN & 0.0754 (0.0011) & \ \ N/A  & 0.664 (0.010) & N/A \\
    \bottomrule
  \end{tabular}
\end{table}

\section{Conclusion}
\label{conclusion}
We have presented a novel approach to modeling highly nonlinear dynamics with a sound quantification of uncertainty, using compositions of random features derived from a first order ordinary differential equation (ODE). Our results show that our model is able to effectively capture nonlinear dynamics in both single and multiple output settings, with the added benefit of competitive performance on benchmark regression datasets. 

Whilst we did find that utilising a kernel based on a first order ODE yielded superior results compared to competing models, the simplicity of the ODE on which our model is based limits its flexibility. However, random features can also be derived from more expressive kernels based on higher order ODEs \citep{guarnizo2018fast} and even partial differential equations. Integrating such kernels into our modeling framework, and perhaps even varying the kernel choice between layers, would be interesting avenues for future research. Another limitation of our approach is that whilst we give $\bm{W}$ and $\bm{\Omega}$ a variational treatment, we do not extend this to the kernel hyperparameters $\bm{\Theta}$; this could also be considered in future work.

In this work, we have demonstrated the applicability of our model to healthcare data in Section \ref{ecg_extrapolation}. Whilst our results demonstrate the capability of our model to learn complex dynamics within a highly nonlinear system, it is important to note that relying heavily on machine learning (ML) software within the domain of healthcare presents the possibility for negative societal impact, specifically ethical and practical issues. Care must be taken to ensure that such software is thoroughly tested and that appropriate safeguards are in place to avoid situations which could cause possible harm to patients. For example, clinical decisions should be made with input from a trained professional, rather than purely based upon information from an ML system.  

\begin{ack}
Thomas M. McDonald thanks the Department of Computer Science at the University of Sheffield for financial support. Mauricio A.~\'Alvarez has been financed by the EPSRC Research Projects EP/R034303/1, EP/T00343X/2 and EP/V029045/1.
\end{ack}

{

\bibliographystyle{unsrtnat}
\bibliography{ms} 

}

\newpage
\appendix

\section{Appendix}

\renewcommand{\thefigure}{A\arabic{figure}}
\setcounter{figure}{0}
\renewcommand{\thetable}{A\arabic{table}}
\setcounter{table}{0}

Within this appendix, we firstly present a table-based comparison of our proposed model with a number of relevant approaches mentioned in the main text. Following this, additional details regarding experimental setups and further justification for our experimental design choices are supplied. We also provide a number of additional experimental results; firstly, we consider an imputation problem using the PhysioNet data from the main paper, before also evaluating deeper variations of the models from the main paper on the PhysioNet extrapolation task and the UCI benchmark datasets. In addition, we evaluate our model on the \textit{FitzHugh-Nagumo} ODE system, the chaotic \textit{Lorenz attractor} system, provide an experimental comparison between our model and \textit{neural processes}, and also include a brief hyperparameter study. We also present a detailed derivation of our variational lower bound, alongside the expression for the Kullback-Leibler divergence between two Gaussian distributions. Finally, we include a short discussion on computational complexity.

\subsection{Model Comparison}
Presented in Table \ref{tab:models_supplemental} is an overview of a number of probabilistic modeling techniques which are relevant to the work we present in this paper.

\begin{table}
  \centering
  \caption{A high-level comparison of the DLFM with a number of other approaches. This includes whether each of the probabilistic models listed is deep or shallow, and whether they incorporate any physical dynamics, followed by a brief summary of the approach.}
    \begin{tabular}{p{0.3\linewidth}ccp{0.3\linewidth}}
    	\toprule
    	Model & Deep? & Physics-Informed? & Summary \\
        \midrule
        LFM \newline \citep{alvarez2009latent} & $\times$ & \checkmark & GP kernel derived from ODEs \\ \midrule
        LFM-RFF \newline \citep{guarnizo2018fast} & $\times$ & \checkmark & Random Fourier features derived from ODEs \\ \midrule
        DLFM (our work) & \checkmark & \checkmark & Random Fourier features derived from ODEs integrated into DGP architecture \\ \midrule
		C-DGP \newline \citep{lorenzi2018constraining} & \checkmark & \checkmark & ODE-based constraints applied to dynamics of a DGP with random Fourier features \\ \midrule
        DGP-RFF \newline \citep{cutajar2017random} & \checkmark & $\times$ & Random Fourier feature expansions applied to a DGP architecture \\ \midrule        
        Deep Hybrid Neural Kernel \newline \citep{mehrkanoon2018deep} & \checkmark & $\times$ & Bridges deep learning and kernel methods using random Fourier features \\ \midrule        
        Deep Factors \newline \citep{wang2019deep} & \checkmark & $\times$ & Probabilistic approach to modeling nonlinear systems which employs GPs and DNNs \\ \midrule 
        IWGP \& SDSP \newline \citep{zammit2021deep} & \checkmark & $\times$ & Modeling of non-stationary data with compositions of stochastic processes \\ \midrule 
        tw-pp-svGPFA \newline \citep{duncker2018temporal} & \checkmark & $\times$ & Nested GP-based model which infers latent time warping functions \\ \midrule
        Conv-CNP \newline \citep{gordon2019convolutional} & \checkmark & $\times$ & An extension of a neural process which is capable of modeling translation equivariance \\ \midrule
        GNP \newline \citep{bruinsma2021gaussian} & \checkmark  & $\times$ & A variant of the Conv-CNP which also models correlations \\
        \bottomrule
    \end{tabular}
  \label{tab:models_supplemental}
\end{table}

\subsection{Experimental Setups}
For all of the experiments performed, we concatenate the output of each layer within the DLFM with the original input to the model as a means of avoiding pathological behaviour \citep{duvenaud2014avoiding}. We also extend this treatment to all of the DGPs tested. The lengthscales and decay parameters within the DLFM are all initialised to 0.01, except at the output layer where we initialise the lengthscales to 1.0. The likelihood variance is also initialised to 0.01, whilst the sensitivity parameters are randomly initialised from a standard normal distribution. We attempted to closely mirror this setup for the DGPs tested by initialising the lengthscales to 0.01, and whilst this lead to improved performance on the dynamical system experiments, such an initialisation resulted in poor performance on the UCI regression benchmarks, therefore we reverted to the initialisation used by the original implementation ($log(D_{F^{(\ell)}})$, where $D_{F^{(\ell)}}$ denotes the dimensionality of the $\ell$-th layer).

\paragraph{Toy Data Experiment} The data was generated by solving the hierarchical ODE system with $\gamma_1 = 0.01$, $\gamma_2 = 0.02$, $\omega=1$, $\tau=0$ and the initial values of $f_1$ and $f_2$ set equal to zero. The extrapolation test data consists of 150 evenly spaced data-points between $t=1.0$ and $t=1.25$, whilst the interpolation test data consists of 100 evenly spaced data-points between approximately $t=0.208$ and $t=0.375$.

\paragraph{PhysioNet Experiments} For the extrapolation experiment, the test data range consisted of the 300 data-points lying between $t=0.7$ and $t=1.0$. For the imputation experiment included in the supplemental material, the test data consisted of 150 contiguous, non-overlapping data-points from each of the three outputs; these were in the range $t=0.20$ to $t=0.35$ for ABP, $t=0.40$ to $t=0.55$ for ECG and $t=0.60$ to $t=0.75$ for ICP.

\paragraph{UCI Experiments} We utilised the same train and test folds as \citet{cutajar2017random} in order to make our results as directly comparable as possible.\footnote{The UCI train and test folds are available at \url{https://github.com/mauriziofilippone/deep_gp_random_features/tree/master/code/FOLDS}.} From each training fold, we set aside 1\% of the observations as a validation set.

\subsection{Additional Experiments}
\subsubsection{PhysioNet Imputation}
\label{ecg_imputation}
\begin{figure}
  \centering
  \includegraphics[width=0.95\linewidth]{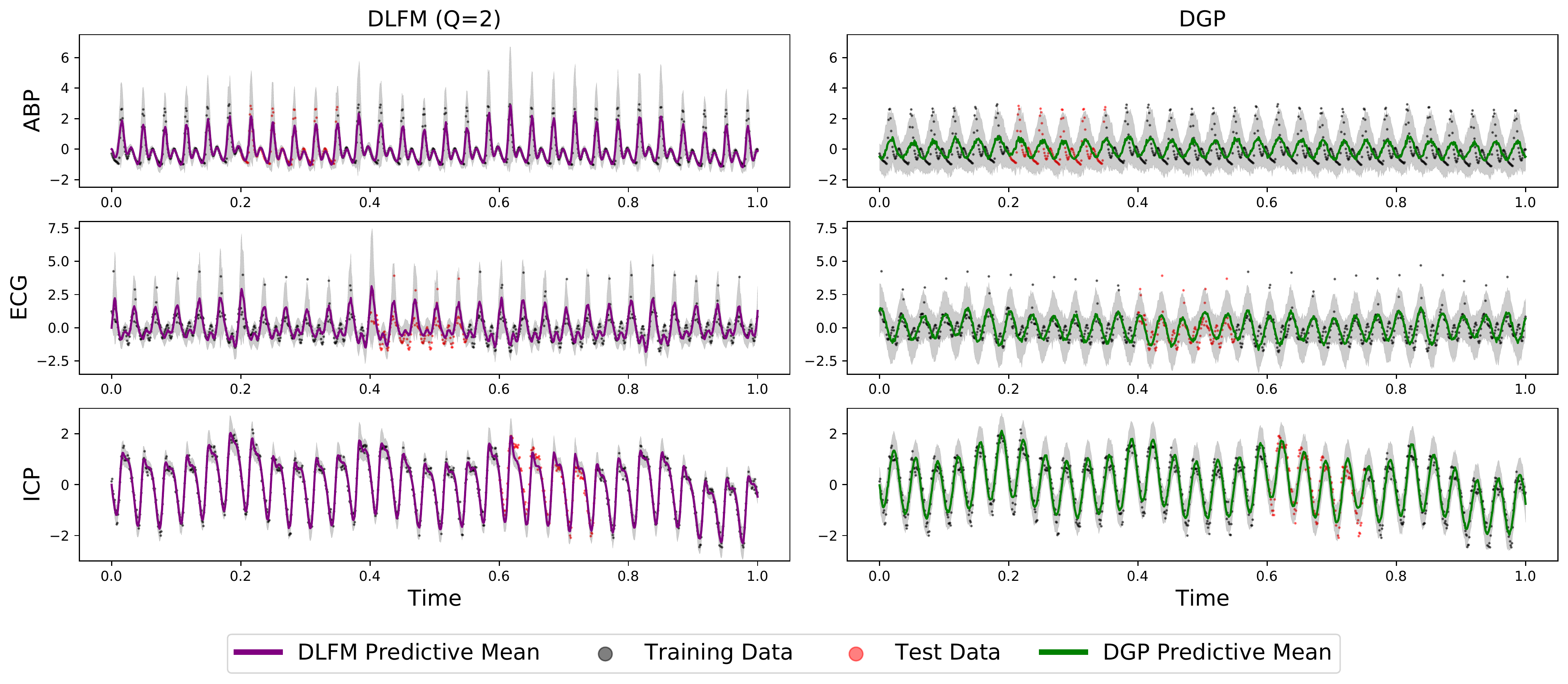}
  \caption{Predictions generated from the imputation scenario described in Section \ref{ecg_imputation} for each of the three outputs within the CHARIS dataset from a DLFM with two latent forces (in the left-hand column), and a DGP with an exponentiated quadratic kernel (in the right-hand column). The shaded grey areas in each plot represent $\pm 2\sigma$.}
  \label{fig:ECG_imputation_figure}
\end{figure}

In this experiment, we test the ability of each model to impute missing output values, by removing 150 non-overlapping data-points from each output during training and using these data-points as a test set. From the results shown in Table \ref{CHARIS_imputation_table} and Figure \ref{fig:ECG_imputation_figure}, we find that although the DGP is capable of imputing the missing outputs with some success, the DLFMs converge to a lower NMSE and MNLL across all three of the outputs. The MNLL metrics for the LFM-RFF are omitted from Table \ref{CHARIS_imputation_table} as the model returned negative predictive variances.

\begin{table}
  \caption{Imputation test set results for each output of the CHARIS dataset, with the standard error reported in brackets.}
  \setlength{\tabcolsep}{2.5pt}
  \label{CHARIS_imputation_table}
  \centering
  \begin{tabular}{ccccccc}
    \toprule &
    \multicolumn{2}{c}{ABP} & \multicolumn{2}{c}{ECG} & \multicolumn{2}{c}{ICP} \\
    \cmidrule(r){2-3} \cmidrule(r){4-5} \cmidrule(r){6-7}
         & NMSE & MNLL & NMSE & MNLL & NMSE & MNLL \\
    \midrule
    DLFM ($Q=1$) & 0.42 (0.06) & 0.79 (0.05) & \textbf{0.48 (0.04)} & 1.00 (0.05) & 0.19 (0.02) & 0.60 (0.04) \\
    DLFM ($Q=2$) & \textbf{0.27 (0.11)} & \textbf{0.49 (0.11)} & 0.68 (0.13) & \textbf{1.05 (0.04)} & \textbf{0.13 (0.01)} & \textbf{0.46 (0.02)} \\
    DGP-EQ & 0.73 (0.02) & 1.18 (0.01) & 0.63 (0.02) & 1.21 (0.01) & 0.23 (0.02) & 0.73 (0.04) \\
    DGP-ARC & 1.00 (0.0003) & 8.10 (0.03) & 1.02 (0.001) & 5.00 (0.02) & 1.03 (0.0004) & 3.60 (0.02) \\
    VAR-GP & 1.00 (0.0001) & 23.9 (13.1) & 1.03 (0.0002) & 10.3 (1.3) & 1.00 (0.0002) & 19.2 (4.1) \\
    DNN & 1.00 (0.001) & N/A & 1.01 (0.001) & N/A & 0.99 (0.001) & N/A \\
    LFM-RFF & 1.00 (0.001) & N/A & 1.01 (0.01) & N/A & 0.97 (0.02) & N/A \\
    \bottomrule
  \end{tabular}
\end{table}

\subsubsection{Additional Hidden Layers}
We carried out two additional experiments to assess the performance of our DLFM with two hidden layers, rather than one. For comparison, we also evaluated two hidden layer variations of the DGP-EQ, DGP-ARC and DNN models mentioned in the main body of the paper. All other aspects of these four models were kept identical to the experimental setups described in the main text.

\begin{table}
  \caption{Extrapolation test set results for the two hidden layer models, for each output within the CHARIS dataset, with standard error in brackets.}
  \setlength{\tabcolsep}{2.5pt}
  \label{CHARIS_table_supplemental}
  \centering
  \begin{tabular}{ccccccc}
    \toprule &
    \multicolumn{2}{c}{ABP} & \multicolumn{2}{c}{ECG} & \multicolumn{2}{c}{ICP} \\
    \cmidrule(r){2-3} \cmidrule(r){4-5} \cmidrule(r){6-7}
         & NMSE & MNLL & NMSE & MNLL & NMSE & MNLL \\
    \midrule
    DLFM ($Q=1$) & \textbf{0.26 (0.08)} & \textbf{0.45 (0.13)} & \textbf{0.55 (0.09)} & \textbf{1.03 (0.08)} & \textbf{0.35 (0.07)} & \textbf{0.92 (0.15)} \\
    DLFM ($Q=2$) & 0.72 (0.21) & 1.79 (0.57) & 0.91 (0.21) & 1.37 (0.19) & 0.54 (0.14) & 1.45 (0.33) \\
    DGP-EQ & 1.01 (0.0001) & 27.1 (1.13) & 1.00 (0.0001) & 27.2 (1.25) & 1.31 (0.0001) & 33.4 (1.41) \\
	DGP-ARC & 1.29 (0.05) & 1.59 (0.01) & 1.27 (0.09) & 1.65 (0.02) & 1.48 (0.03) & 1.61 (0.01) \\
	DNN & 1.01 (0.001) & N/A & 1.02 (0.01) & N/A & 1.35 (0.01) & N/A \\
    \bottomrule
  \end{tabular}
\end{table}

\paragraph{PhysioNet Multivariate Time Series} Considering once again the extrapolation problem described in the main paper, Table \ref{CHARIS_table_supplemental} contains the test set NMSE and MNLL corresponding to each output, for each of the models mentioned above, with associated standard error reported with respect to the random seed. The results from evaluating the DLFMs are largely similar to the single hidden layer case, with the $Q=1$ variant continuing to outperform all of the other techniques tested. We found that regardless of the scale of the initial lengthscales chosen, the two hidden layer DGPs struggled to converge during training, leading to much higher NMSE and MNLLs than those achieved by the single hidden layer DGPs in the main paper.

\begin{table}
  \caption{Test set results on the UCI benchmarks for the two hidden layer models, with standard error in brackets.}
  \label{UCI_table_supplemental}
  \centering
  \begin{tabular}{ccccc}
    \toprule &
    \multicolumn{2}{c}{Powerplant} & \multicolumn{2}{c}{Protein} \\
    \cmidrule(r){2-3} \cmidrule(r){4-5}
         & NMSE & MNLL & NMSE & MNLL \\
    \midrule
    DLFM & 0.0563 (0.001) & -0.0237 (0.01) & 0.706 (0.003) & 0.7335 (0.01) \\
    DGP-EQ & \textbf{0.0558 (0.001)} & 0.0677 (0.06) & \textbf{0.574 (0.009)} & \textbf{0.5678 (0.03)}  \\
    DGP-ARC & 0.0612 (0.001) & \textbf{-0.0731 (0.01)} & 0.680 (0.030) & 0.9308 (0.01)  \\
    DNN & 0.0965 (0.005) & N/A & 0.614 (0.005) & N/A \\
    \bottomrule
  \end{tabular}
\end{table}

\paragraph{UCI Regression Benchmarks} Considering the two UCI regression benchmark datasets once again, Table \ref{UCI_table_supplemental} contains the test set NMSE and MNLL for each of the four models with two hidden layers, averaged over three folds, with associated standard error reported with respect to the random seed. The inclusion of an additional hidden layer in the DLFM leads to broadly similar results to those seen for the single hidden layer case. The DGPs also tend to to follow this trend, however they converge to a considerably lower MNLL than their single hidden layer counterparts. The single hidden layer DLFM does still marginally outperform the two hidden layer DGP-EQ in terms of NMSE, and the shallow VAR-GP model from the main paper converges to a lower NMSE on the protein dataset than all of the two hidden layer models tested.

\subsubsection{FitzHugh-Nagumo System}
As discussed in the main paper, the physically constrained DGP (C-DGP) of \cite{lorenzi2018constraining} is not directly applicable to problems such as the PhysioNet experiments, as the C-DGP requires that the ODE system driving the observed dynamics must be known. However, we can compare our DLFM to the C-DGP by evaluating the predictive performance of both models on a scenario in which the ODE system is known, such as the \textit{FitzHugh-Nagumo} system \citep{fitzhugh1955mathematical}. Specifically, we consider the experimental setup used by \cite{lorenzi2018constraining}, which consists of 400 noisy observations of the ODE system. Firstly, we compare the performance of single hidden layer versions of both our DLFM, the C-DGP and the previously mentioned DGP-EQ on these noisy observations and evaluate the NMSE between the model predictions and the underlying ground truth, which is the exact solution of the ODE system with no noise added. These results are shown under Scenario A in Table \ref{tab:FitzHugh}. The results reported in Scenario B of Table \ref{tab:FitzHugh} correspond to an alternate scenario in which we test the ability of all three models to extrapolate the dynamics by training on the first 300 observations and evaluating the NMSE on the final 100 observations.

\begin{table}
  \centering
  \caption{Test set NMSE values for both FitzHugh-Nagumo experiments.}
    \begin{tabular}{ccc}
    	\toprule
    	 & Scenario A - Full Data Fit & Scenario B - Extrapolation Task \\
        \midrule
        DLFM & 0.02 & \textbf{0.07}  \\
        C-DGP & \textbf{0.01} & 0.38  \\
        DGP-EQ & 0.03 & 0.12  \\
        \bottomrule
    \end{tabular}
  \label{tab:FitzHugh}
\end{table}

From the results shown in Table \ref{tab:FitzHugh}, we can see that the C-DGP slightly outperforms the DLFM on Scenario A, likely because the C-DGP explicitly includes constraints specific to this system. However, the DLFM has a greater capacity to accurately extrapolate the dynamics despite not having any specific prior knowledge of the system, as evidenced by the results for Scenario B. The DLFM also outperforms the DGP-EQ in both tasks. The DGP-EQ tested had a single hidden layer of dimensionality 3 and 100 random features and the DLFM tested had an identical architecture but used 2 latent forces with 50 random features per latent force, for parity. The C-DGP tested had an identical architecture to that used by \cite{lorenzi2018constraining} in their experiments on this system.

\subsubsection{Lorenz Attractor}
We also conducted a brief additional experiment on a chaotic dynamical system, the \textit{Lorenz attractor} \citep{lorenz1963deterministic},

\begin{align*}
\frac{dx}{dt} &= \sigma(y-x) \\
\frac{dy}{dt} &= x(r-z)-y \\
\frac{dz}{dt} &= x y-b z
\end{align*}

Utilising the commonly used system parameters $\sigma = 10$, $b = 8 /3$ and $r = 28$ which result in chaotic behaviour, we generated 1000 data-points from this system in the interval $t=0$ to $t=50$. Two different modeling scenarios were tested, ‘80:20’ where the first 80\% of the data was used for training and the final 20\% withheld for testing, and a ‘98:2’ split where only the final 2\% of the time series was withheld for testing. DLFM and DGP-EQ models identical to those used in the FitzHugh-Nagumo experiment were employed, however we utilised two hidden layers rather than one, as both models seemed to benefit from this added depth. The test set NMSE values are shown in Table \ref{tab:Lorenz_supplemental}.

\begin{table}
  \centering
  \caption{Test set NMSE values for two hidden layer models evaluated on the Lorenz attractor system.}
    \begin{tabular}{ccc}
    	\toprule
    	 & 80:20 Split & 98:2 Split \\
        \midrule
        DLFM & \textbf{0.94} & \textbf{0.89} \\
        DGP-EQ & 1.09 & 1.24 \\
        \bottomrule
    \end{tabular}
  \label{tab:Lorenz_supplemental}
\end{table}

We can see from these brief results presented in Table \ref{tab:Lorenz_supplemental} that the DLFM outperforms the DGP-EQ in both scenarios, however a more rigorous series of experiments would need to be performed to accurately assess the ability of our model to model chaotic dynamical systems.

\subsubsection{Neural Process Comparison}
\textit{Neural processes} (NPs) are a recently proposed class of model, developed in order to combine a number of the benefits of neural networks and Gaussian processes \citep{garnelo2018neural}. Subsequent research in this area has focused on applications to time-series modeling by modeling \textit{translation equivariance} in data; the \textit{convolutional conditional neural process} (Conv-CNP) \citep{gordon2019convolutional} and the \textit{Gaussian neural process} (GNP) \citep{bruinsma2021gaussian} are two examples of such work.

Given that time-series modeling is a key application for our proposed model, we present a brief experimental comparison between our approach and the Conv-CNP and GNP frameworks. Specifically, we considered a revised version of the PhysioNet extrapolation problem from the main paper. Following the approach taken in Section 5.1 of the work of \cite{gordon2019convolutional}, we evaluate the test set log likelihood on 1-D time series, training and evaluating a separate model for each output of the PhysioNet CHARIS dataset in turn. However, we consider a more challenging split of the training and test set data, with the first 500 data-points used for training and the final 500 withheld for testing. The deep LFM used for this experiment consisted of a single hidden layer with one latent force and 100 random Fourier features. The test set MNLL values obtained are shown in Table \ref{tab:NP_supplemental}.

\begin{table}
  \centering
  \caption{Test set MNLL values for the DLFM and NP models trained and evaluated on each individual output of the PhysioNet dataset in turn.}
    \begin{tabular}{cccc}
    	\toprule
    	 & ABP & ECG & ICP \\
        \midrule
        DLFM & 1.53 & 1.61 & \textbf{1.51} \\
        Conv-CNP & 1.59 & \textbf{1.47} & 1.84 \\
        GNP & \textbf{1.50} & 1.50 & 1.66 \\
        \bottomrule
    \end{tabular}
  \label{tab:NP_supplemental}
\end{table}

From the results shown in Table \ref{tab:NP_supplemental}, it is clear that the DLFM achieves competitive results compared to the Conv-CNP, outperforming the Conv-CNP on the ABP and ICP outputs. The GNP outperforms the DLFM on two of the outputs, but as with the DLFM comparison to the Conv-CNP, the results are relatively similar, and further analysis would be required to fully assess how the models compare under different conditions, and how their performance compares to other deep models.

\subsubsection{Hyperparameter Testing}
Selecting the optimal architecture and hyperparameters for a given model is in itself a challenging research problem across sub-fields of deep learning concerned with both deterministic and probabilistic modeling. Whilst we do not provide an exhaustive assessment of the impact of certain hyperparameters on the performance of our model (primarily because this will very much depend on the data being considered), we present a brief hyperparameter study on the PhysioNet extrapolation problem from the main paper, which illustrates the impact of changing certain aspects of the DLFM. The results of this study are shown in Table \ref{tab:HP_supplemental}, with the metrics averaged over all three outputs.

\begin{table}
  \caption{Test set results averaged across all three outputs on the PhysioNet extrapolation experiment for a range of DLFM architectures. $D_F$ denotes the dimensionality of the hidden layer, $N_{RF}$ denotes the number of random Fourier features, and $Q$ denotes the number of latent forces used.}
  \label{tab:HP_supplemental}
  \centering
  \begin{tabular}{ccccccccc}
    \toprule 
    $D_F$ &
    \multicolumn{4}{c}{1} & \multicolumn{4}{c}{6} \\
    \cmidrule(r){2-5} \cmidrule(r){6-9}
    $N_{RF}$ & \multicolumn{2}{c}{10} & \multicolumn{2}{c}{50} & \multicolumn{2}{c}{10} & \multicolumn{2}{c}{50} \\
    \cmidrule(r){2-3} \cmidrule(r){4-5} \cmidrule(r){6-7} \cmidrule(r){8-9}
       $Q$ & 1 & 6 & 1 & 6 & 1 & 6 & 1 & 6 \\
    \midrule
    NMSE & 0.90 & 0.60 & 0.66 & 0.56 & 0.30 & 0.34 & \textbf{0.26} & 0.54 \\
    MNLL & 1.52 & 1.44 & 1.41 & \textbf{1.11} & 3.34 & 3.95 & 2.02 & 1.19 \\

    \bottomrule
  \end{tabular}
\end{table}

We can see from Table \ref{tab:HP_supplemental} that increasing the number of random Fourier features appears to reduce the predictive error and also significantly improves uncertainty quantification, as shown by the reduced MNLL values. Increasing the width of the hidden layer also broadly tends to improve performance.

\subsection{Derivation of the Variational Lower Bound}
Denoting $\bm{\Psi} = \left\{\bm{W}, \bm{\Omega} \right\}$ for ease of notation, the variational lower bound on the marginal likelihood can be derived as follows:
\begin{align*}
\log[p(\bm{y} | \bm{X}, \bm{\Theta})] &= \log \left[\int p(\bm{y} | \bm{X}, \bm{\Psi}, \bm{\Theta}) p(\bm{\Psi} | \bm{\Theta}) d\bm{\Psi} \right] \\ 
&= \log \left[\int \frac{p(\bm{y} | \bm{X}, \bm{\Psi}, \bm{\Theta}) p(\bm{\Psi} | \bm{\Theta})}{q(\bm{\Psi})} q(\bm{\Psi}) d\bm{\Psi} \right] \\
&= \log \left[E_{q(\bm{\Psi})} \frac{p(\bm{y} | \bm{X}, \bm{\Psi}, \bm{\Theta}) p(\bm{\Psi} | \bm{\Theta})}{q(\bm{\Psi})} \right] \\
&\geq E_{q(\bm{\Psi})} \left( \log \left[ p(\bm{y} | \bm{X}, \bm{\Psi}, \bm{\Theta}) \right] \right) + E_{q(\bm{\Psi})} \left(\log \left[\frac{p(\bm{\Psi} | \bm{\Theta})}{q(\bm{\Psi})}\right] \right) \\
&= E_{q(\bm{\Psi})} \left( \log[p(\bm{y} | \bm{X, \Psi, \Theta})] \right) - \text{DKL}[q(\bm{\Psi}) || p(\bm{\Psi} | \bm{\Theta})] \\
&\approx \left[\frac{N}{M} \sum_{k \in \mathcal{I}_M} \frac{1}{N_{\text{MC}}} \sum^{N_{\text{MC}}}_{r=1} \log[p(\bm{y}_k | \bm{x}_k, \tilde{\bm{\Psi}}_r, \bm{\Theta})]\right] - \text{D}_{\text{KL}}[q(\bm{\Psi})||p(\bm{\Psi} | \bm{\Theta})] .
\end{align*}
We assume a factorised prior over the spectral frequencies and weights across all layers, which takes the form,
\begin{align*}
p(\bm{\Psi}|\bm{\Theta}) = \prod_{\ell=0}^{L-1} p\left(\bm{\Omega}^{(\ell)} | \bm{\Theta}^{(\ell)}\right)  p\left(\bm{W}^{(\ell)}\right) = \prod_{ij\ell} q\left(\Omega_{ij}^{(\ell)}\right) \prod_{ij\ell} q\left(W_{ij}^{(\ell)}\right) ,
\end{align*}
where, 
\begin{align*}
q\left(\Omega_{ij}^{(\ell)} \;\middle|\; m_{ij}^{(\ell)}, (s^2)_{ij}^{(\ell)} \right) &= \mathcal{N}\left(\Omega_{ij}^{(\ell)} \;\middle|\; m_{ij}^{(\ell)}, (s^2)_{ij}^{(\ell)}\right) \\
q\left(W_{ij}^{(\ell)} \;\middle|\; \mu_{ij}^{(\ell)}, (\beta^2)_{ij}^{(\ell)} \right) &= \mathcal{N}\left(W_{ij}^{(\ell)} \;\middle|\; \mu_{ij}^{(\ell)}, (\beta^2)_{ij}^{(\ell)}\right).
\end{align*} 
As discussed in the main paper, we reparameterize the spectral frequencies and weights as,
\begin{align*}
\left(\tilde{\Omega}_r^{(\ell)}\right)_{ij} = s_{ij}^{(\ell)} \epsilon_{rij}^{(\ell)} + m_{ij}^{(\ell)} \\
\left(\tilde{W}_r^{(\ell)}\right)_{ij} = \beta_{ij}^{(\ell)} \epsilon_{rij}^{(\ell)} + \mu_{ij}^{(\ell)} 
\end{align*}
where $\epsilon_{rij}^{(\ell)}$ represent samples from a standard normal, $\tilde{\Omega}_r \sim q\left(\Omega_{ij}^{(\ell)}\right)$ and $\tilde{W}_r \sim q\left(W_{ij}^{(\ell)}\right)$. We can then optimise our lower bound with respect to the parameters governing our variational distributions $\left(m_{ij}^{(\ell)}, (s^2)_{ij}^{(\ell)}, \mu_{ij}^{(\ell)} \text{ and } (\beta^2)_{ij}^{(\ell)}\right)$ and our kernel hyperparameters $\bm{\Theta}$, using conventional gradient descent techniques.

\subsubsection{KL Divergence Between Normal Distributions}
The Kullback-Leibler (KL) divergence between two normal distributions $p_A\left(x \;\middle|\; \mu_A, \sigma_A^2 \right) = \mathcal{N}\left(x \;\middle|\; \mu_A, \sigma_A^2 \right)$ and $p_B\left(x \;\middle|\; \mu_B, \sigma_B^2 \right) = \mathcal{N}\left(x \;\middle|\; \mu_B, \sigma_B^2 \right)$ can be expressed by,

\begin{align*}
\text{DKL}\left[p_A(x) || p_B(x) \right] = \frac{1}{2} \left[\log \left(\frac{\sigma_B^2}{\sigma_A^2} \right) -1 + \frac{\sigma_A^2}{\sigma_B^2} + \frac{(\mu_A - \mu_B)^2}{\sigma_B^2} \right] .
\end{align*}

\subsection{Computational Complexity}
The computational complexity associated with our stochastic variational inference scheme is $\mathcal{O}(m D Q N_{RF} N_{MC})$ where $m$ denotes the mini-batch size, $D$ denotes the layer dimensionality, $Q$ and $N_{RF}$ represent the number of latent forces and random Fourier features respectively and $N_{MC}$ denotes the number of Monte Carlo samples used. This very closely follows the computational complexity of the DGP with random feature expansions presented by \cite{cutajar2017random}, with an added linear dependency on $Q$, the number of latent forces employed.

\end{document}